\documentclass{IOS-Book-Article}

\usepackage{mathptmx}
\usepackage{graphicx}
\usepackage{svg}
\usepackage{amsmath} 
\usepackage{mwe}

\usepackage{hyperref}
\usepackage{todonotes}

\usepackage{xcolor}
\usepackage{tabularx}
\usepackage{lastpage}
\def\hb{\hbox to 10.7 cm{}}

\begin{document}

\pagestyle{headings}
\def\thepage{}

%\pagenumbering{arabic}

\begin{frontmatter}              % The preamble begins here.

%\pretitle{Pretitle}
\title{Legal Search in Case Law \\and Statute Law}

\markboth{}{December 2019\hb}

%\subtitle{Number of pages: \pageref{LastPage}}
%\subtitle{Double submission to JURISIN 2019 and JURIX 2019}

\author[A]{\fnms{Julien} \snm{Rossi}%
\thanks{Corresponding Author: \href{mailto:j.rossi@uva.nl}{j.rossi@uva.nl}}},
\author[A,B]{\fnms{Evangelos} \snm{Kanoulas}}

\runningauthor{J. Rossi and E. Kanoulas}
\address[A]{Amsterdam Business School, University of Amsterdam}
\address[B]{Institute of Informatics, University of Amsterdam}

\begin{abstract}
In this work we describe a method to identify document pairwise relevance in the context of a typical legal document collection: limited resources, long queries and long documents. We review the usage of generalized language models, including supervised and unsupervised learning. We observe how our method, while using text summaries, overperforms existing baselines based on full text, and motivate potential improvement directions for future work.
\end{abstract}

\begin{keyword}
Legal Search \sep Case Law \sep Statute Law \sep Language Models \sep Transfer Learning
\end{keyword}

\end{frontmatter}

\markboth{December 2019\hb}{December 2019\hb}

\section{Introduction}
The increasing amount of legal data available requires the ability to search and identify relevant information in this data, it calls for the ability to automate or assist in specialized retrieval tasks, as a service to either the public or practitioners. Legal Search is the specialized Information Retrieval (IR) task dealing with legal information relevant to a situation described factually and legally. The legal documents collections differ from generic IR collections by the level of professional knowledge needed to produce a labeled corpus~\cite{arora2018challenges}. This results in limited and specialized collections, each covering a narrow field of interest: lease contracts, financial products, court decisions, etc\ldots Existing literature shows how Legal Search is an essential tool for practitioners and citizens in need of critical information~\cite{nejadgholi2017semi, maclean2015delivering, boyd2015alienated}, whereas it introduces difficult challenges that our work addresses.

In this paper, we focus on limitations of the applicability of generalized language models~\cite{radford2019language, DBLP:journals/corr/abs-1801-06146, DBLP:journals/corr/abs-1810-04805, GeneralizedLM} in the context of the handling of legal information tasks. We observe that specific features of typical legal documents adversely affects those models, such as long texts, mixing of formal abstract concepts with casual verbatims in layman language, or dealing with an unclear formulation of relevance.  

The development of Generalized Language Models such as BERT~\cite{DBLP:journals/corr/abs-1810-04805} has allowed the application of complex neural models to tasks even with limited data available, significantly improving over lexical methods, while past neural models were having difficulties to deal with smaller datasets, due to vanishing gradients, and were difficult to compute and implement on hardware~\cite{StopRNN}.

The core of these generalized language models is the separation between learning the language (pre-training) and adapting to a task (fine-tuning). The adaptation to legal tasks is complicated by the difference between the generic language and the legalese language, a difference in word usage and meaning, information structuring, or sentence complexity~\cite{legallanguage}.

Our contribution is a method to improve Legal Information Retrieval, where the ranking problem is reformulated as a pairwise relevance score problem, modeled as a fine-tuning task of a generalized language model, and where we adapt long documents to limited input sequence length by summarizing. Furthermore we show how additional pre-training based on a narrow selection of texts can improve performance. 

We will focus on the following questions:
\begin{itemize}
    \item \textbf{RQ1} \textit{Can we use summary encoding as a dense representation of long documents~?}
    \item \textbf{RQ2} \textit{Can we use pre-training and fine-tuning of neural language model with limited data to learn a specific legal language~?}
\end{itemize}

In section~\ref{section_methods}, we will introduce our methods to answer the research questions, then in section~\ref{section_task} the tasks we will use for our work, then the research for each task will be described with results and analysis in sections~\ref{section_case_law} and \ref{section_statute_retrieval}. We will present our global analysis and conclusion in section~\ref{section_conclusion}, along with motivation for future research.

\section{Methods} \label{section_methods}
Our work reformulates the ranking problem into a pairwise relevance classification problem, where we train a classifier to separate relevant pairs (made of a query case and a noticed case) from irrelevant pairs of documents. For each query, we let the model infer the probability for the positive class for each pair made of the query and a candidate document, and then rank the candidate documents according to that score. 

This method follows after the well-established probability ranking principle in IR~\cite{robertson1977probability}. For the tasks we consider, we argue that the relevance of a document with regards to a given query is independent from other documents of the collection, confirming the applicability of this framework to our tasks. 

Within this framework, we introduce input pre-processing as well as model training steps that will help us address the research questions. We refer to~\cite{DBLP:journals/corr/abs-1810-04805} for details on the topology of BERT as well as for the setup of the pre-training task or the fine-tuning task.

\subsection{Pairwise Embeddings}
We make use of the capacity of BERT models to encode single dense embeddings for a pair of texts. These embeddings encode the interaction between the texts in the pair, so that it is a suitable input for a downstream pair classifier. In a BERT system, this corresponds to the embeddings of \texttt{[CLS]} token, inserted at the beginning of each input sequence.

\subsection{Fine-Tuning of Pre-Trained Language Model}
The fine-tuning is a supervised learning task for a complete model comprising a pre-trained BERT and a fully-connected layer used for classification. The learning is driven by the true classification labels of the examples, and all the weights of the combined model are updated during this training. The usual loss for classification is the Cross-Entropy. During this phase, the combined model learns how to perform the classification based on the pairwise embeddings.

\subsection{In-Domain Additional Pre-Training of Language Model}
This task is an unsupervised learning task, where an already pre-trained BERT model receives further pre-training with additional texts. We make the remark that BERT is a NLP system, with a high capacity of understanding at the language level. For example~\cite{Lin2019,Clark2019} demonstrate how the layers in BERT work as a hierarchical system with a capacity to discover syntactic features of the text, while the semantic features are assumed to emerge naturally from co-location and word contexts. In that regard, legal texts create a variety of domain specific languages, considering different domains such as contracts, court decisions, legislation, etc. 

Legal languages differ from casual languages with unusual vocabulary (rare words, latin words, etc.), semantics (words where the legal meaning differs from casual usage), and syntactic features. We suggest that a pre-trained model would learn specific knowledge from a limited additional pre-training on legal domain-specific texts. We base our suggestion on the fact that within a specific legal domain, the text features will be fairly uniform. Following~\cite{ruder2019transfer}, we expect pre-training on the available material being more beneficial to the capacities of the system than only fine-tuning for the task. We expect to answer RQ2 with this method.

\subsection{Summarization of Long Documents}
This method will allow us to address our question RQ1, on long documents. As we base our models on BERT, we are constrained by a maximum input length of 512 WordPiece~\cite{DBLP:journals/corr/WuSCLNMKCGMKSJL16} tokens. WordPiece will subdivide each word into multiple tokens, we consider that this will nearly double the number of tokens observed under a standard Punkt~\cite{Kiss:2006:UMS:1245119.1245122} tokenizer as implemented in NLTK~\cite{BirdKleinLoper09}. 

For this reason, we introduce an extractive summarization of the texts in the corpus using TextRank~\cite{DBLP:journals/corr/BarriosLAW16}. TextRank is an extractive summarization model, based on a graph ranking model operating at sentence level. As it extracts full sentences, it preserves the structure of a natural language and makes the summary fit for input to a Neural Language Model. We choose to limit the size of the summary to 180 words, so a concatenated pair of texts will not be longer than 512 WordPiece tokens. In the case this limit of 512 is not respected, then the sequence will drop the last tokens from the text of candidate case. We use the implementation from gensim~\cite{rehurek_lrec}.

\section{Legal Retrieval Tasks} \label{section_task}
We will illustrate our work with 2 different tasks and their associated document collection, taken from COLIEE~\cite{coliee}. The 2 tasks illustrate different aspects of our research:
\begin{itemize}
    \item Case Law Retrieval: Find past cases similar to a query case. In this task, the query and the documents are both long texts, and the amount of available data is limited. We will address RQ1 and RQ2 with this task.
    \item Statute Law Retrieval: Find law articles relevant to a situation. In this task, the query and documents are short snippets of text, and only a small amount of labeled data is available. We will address RQ2 with this task.
\end{itemize}

\subsection{Case Law Retrieval} \label{subsection_case_law}
In this task, a given court case is considered as a query to retrieve supporting cases (also named 'noticed cases') for the query case. A noticed case supports the decision taken in the query case, although the final decision itself is irrelevant in our retrieval. What matters is the proximity of the themes that are tackled by the query case and the noticed cases, either legal themes or narrative themes.

The dataset is drawn from an existing collection of Federal Court of Canada case law, provided by vLex Canada\footnote{\url{http://ca.vlex.com}}. Each query case is given a collection of 200 potential supporting cases (also named 'candidate cases'), these collections are provided labeled for the training dataset, and unlabelled for the unknown test dataset. The training dataset contains 285 query cases. All cases are relative to Citizenship and Immigration proceedings.

\begin{figure}[ht]
    \centering
    \begin{minipage}{0.49\columnwidth}
        \centering
        \includegraphics[width=\textwidth]{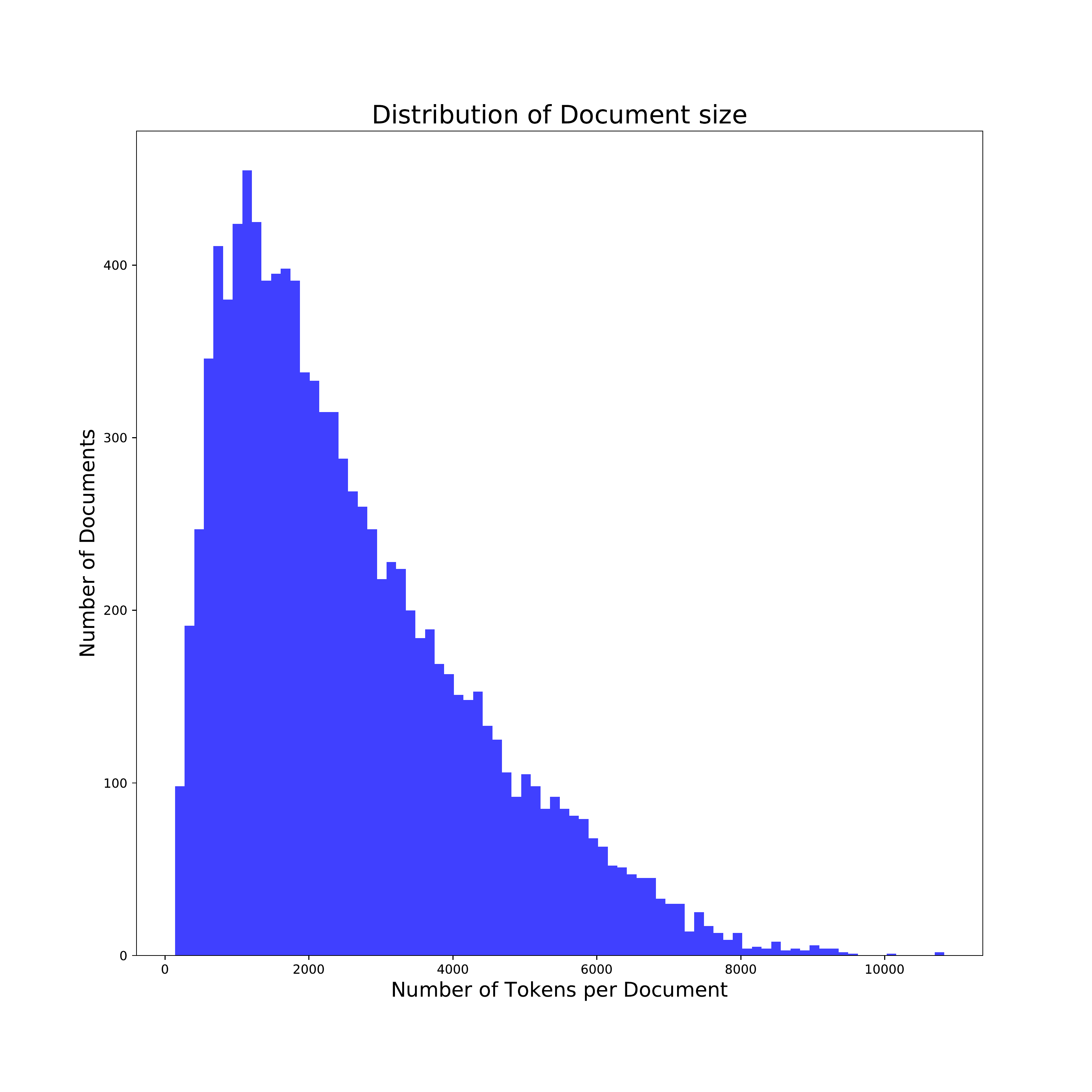}
        \caption{Distribution of the number of words per document}
        \label{pdf_tokens}
    \end{minipage}\hfill
    \begin{minipage}{0.02\columnwidth}
    
    \end{minipage}
    \begin{minipage}{0.49\columnwidth}
        \centering
        \includegraphics[width=\textwidth]{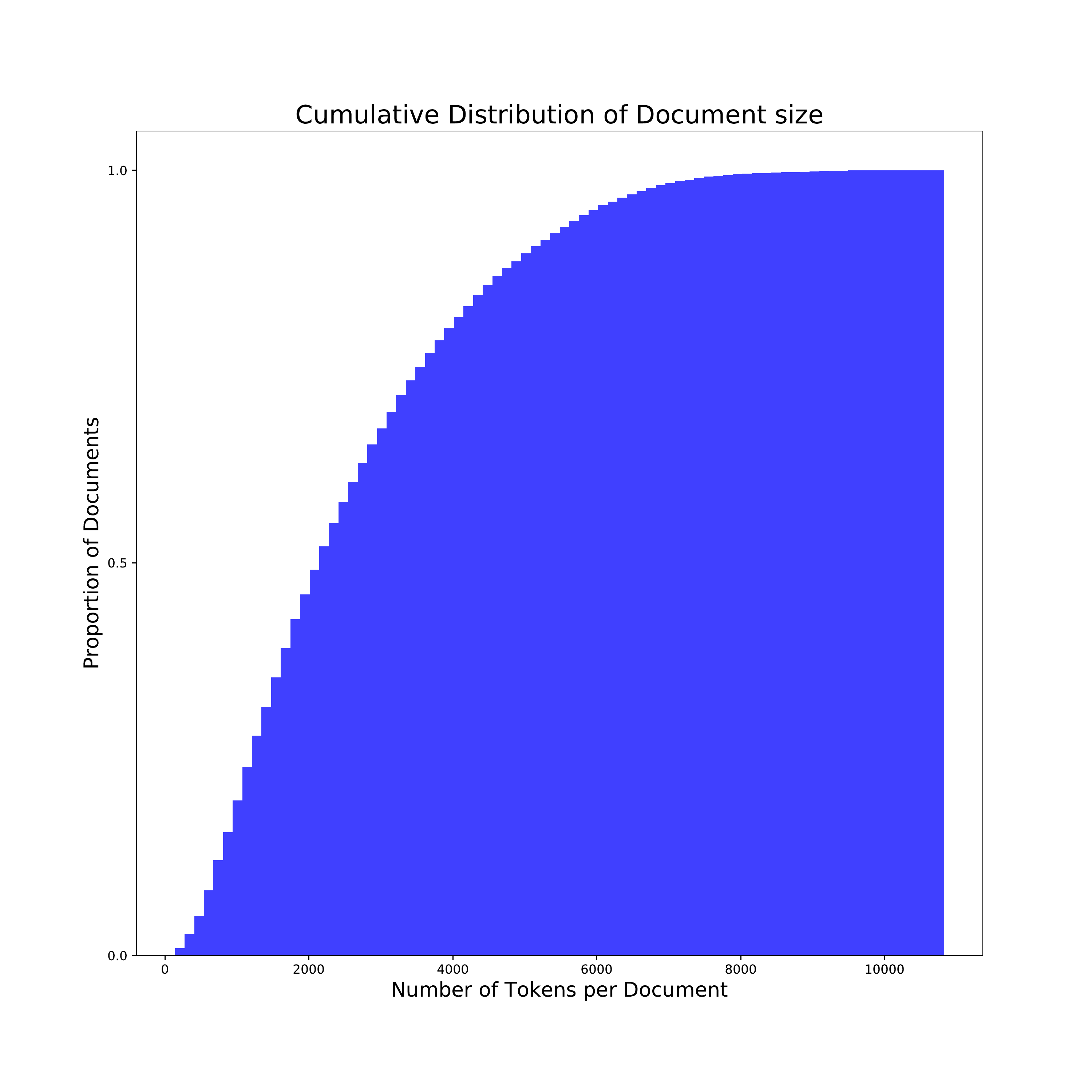}
        \caption{Cumulative distribution for the number of words per document}
        \label{cdf_tokens}
    \end{minipage}
 \end{figure}

We consider the corpus as the complete collection of query cases, and candidate cases from the training dataset, Figures~\ref{pdf_tokens} and \ref{cdf_tokens} illustrate the distribution of document length across the corpus, measured in number of words. With a median length of 2500 words, this dataset illustrates well the challenge of dealing with long documents and long queries, as formulated in our RQ1. The corpus is a limited resource with only 285 queries, which fits well with our RQ2.

\subsection{Statute Law Retrieval} \label{subsection_statute_law_retrieval}
In this task, a brief query has to be disputed with regards to the existing legislation, in this case limited to the Japanese Civil Code. The retrieval task identifies which articles of the Civil Code are relevant to the evaluation of the legal validity of the query.  

The data for this task is compiled from Japanese Bar Exams, and provided both in Japanese and English. We focus on the English dataset from the 2018 edition. The training dataset contains 651 queries, while the test dataset is made of 69 queries. This amount of data will be a good fit for our RQ2. The queries are short snippets with an average of 40 words, with a maximum of under 120 words, while the average law article has 60 words, with 98\% of articles having less than 200 words. We will not face the issue of long texts. We consider it a limited resource with few examples of relevant matches, as 94\% of queries have 1 or 2 relevant articles, among around 1000 candidate articles.

\subsection{Evaluation Metrics} \label{subsection_evaluation_metrics}
In the COLIEE setting, the system decides for each query the number of candidates it returns, and it is evaluated based on the F1 or F2 score of that sub-list. In this paper, we will consider a more traditional approach with respect to Information Retrieval, grounded in a realistic use case for the Case Law Retrieval Task: the end user is a staff from a judge's office, and the queries are submitted to a computer software and returned as a ranked list. We use the same metrics for the Statute Law Retrieval Task.

We choose to report Precision at R (P@R), where R is the total number of relevant documents, Recall and Precision at k (R@k, P@k) and Mean Average Precision (MAP). For each metric, we will consider the macro average\footnote{\(m_i\) is the value of the metric for sample \(i\in[1,N]\), then \(macro(m)=\frac{1}{N}\sum_{i=1}^{N}m_i\)}. Metrics are computed with \texttt{trec\_eval}\footnote{\url{https://github.com/usnistgov/trec\_eval}}, and Precision-Recall curve is plotted with \texttt{plot-trec\_eval}\footnote{\url{https://github.com/hscells/plot-trec\_eval}}

\section{Case Law Retrieval} \label{section_case_law}
\subsection{Experimental Setup} \label{setup_01}
Following the "Methods" section~\ref{section_methods}, we pre-process the collection by summarizing all texts. The final dataset is a collection of triplets made of query text, candidate text and relevance judgment. We will fine-tune a pre-trained BERT model for a binary classification task, using the relevance labels from the labeled training dataset. We will also perform additional in-domain pre-training, and then fine-tune this model for the same classification task. In the results section~\ref{subsection_case_law_results}, the models appear under the names \texttt{FineTuned} and \texttt{PreTrained}.

For the in-domain pre-training, we use the entire corpus of court decisions as the pre-training corpus. This corpus has 18000 documents, and a total of 45 million tokens. The training will start with a pre-trained BERT model (bert-base-uncased), using the \texttt{Megatron-LM}\footnote{\url{https://github.com/cybertronai/Megatron-LM}} library. We run the pre-training on 1 GPU nVidia Tesla P40, with 24GB onboard RAM during 24 hours, or 70000 iteration steps. We export this model for fine-tuning with \texttt{fast-bert}\footnote{\url{https://github.com/kaushaltrivedi/fast-bert}}, during 10 hours for 4 epochs.

The labeled dataset is split 75\%/25\% between training data and evaluation data. The dataset is split according to the cases, so that the cases in the evaluation dataset are not in the training dataset. This split reflects properly the capacity of the system to generalize to unseen data. In this setting, there will also be candidate cases that are only in the evaluation dataset, and therefore unseen data.

For both models, after the fine-tuning step is finalized, the trained model computes the pairwise relevance score for each pair of query case and candidate case in the test dataset, the score for the positive class is used to rank the candidate cases of each query case.

Our implementation is based on libraries \texttt{pytorch-transformers}\footnote{\url{https://github.com/huggingface/pytorch-transformers}}, using \texttt{PyTorch}\footnote{\url{https://pytorch.org/}} framework.

\subsection{Baselines} \label{section_baselines}
We consider 2 lexical features baselines based on BM25. As we setup our BERT-based models to learn from the summaries of the full documents, we propose to have a baseline that represents as well this limitation in the availability of information:
\begin{itemize}
    \item One baseline will use BM25 score as the pairwise relevance score, when the corpus contains all full texts documents
    \item Another baseline will use BM25 score as the pairwise relevance score, with a corpus made of the summaries of all documents
\end{itemize}

Further in the Results section, we will also introduce the Perfect Ranker, this ideal system ranks first all noticed cases, and then all unnoticed cases. It will provide us with an upper bound for the performances at each rank.

\subsection{Results and Analysis} \label{subsection_case_law_results}
\begin{table}[ht]
\begin{tabularx}{\linewidth}{|X|rr|rr|rr|}
    \hline
    System        & R@10 & P@10 & R@1 & P@1  & P@R  & MAP \\
    \hline
    BM25 Summaries    & 0.14 & 0.07 & 0.02 & 0.11 & 0.11 & 0.14 \\
    BM25        & 0.76 & 0.36 & 0.32 & 0.70 & 0.68 & 0.73 \\
    PERFECT     & \textit{0.97} & \textit{0.50} & \textit{0.41} & \textit{1.00} & \textit{1.00} & \textit{1.00} \\
    \hline
    Fine-Tuned   & 0.75 & 0.34 & 0.31 & 0.80 & 0.63 & 0.70 \\
    Pre-Trained & \textbf{0.81} & \textbf{0.39} & \textbf{0.34} & \textbf{0.90} & \textbf{0.73} & \textbf{0.79} \\
    \hline
\end{tabularx}
\caption{Table of Results for All Systems}
\label{01_results_all}
\end{table}

\begin{figure}[ht]
  \centering
  \includegraphics[width=\linewidth]{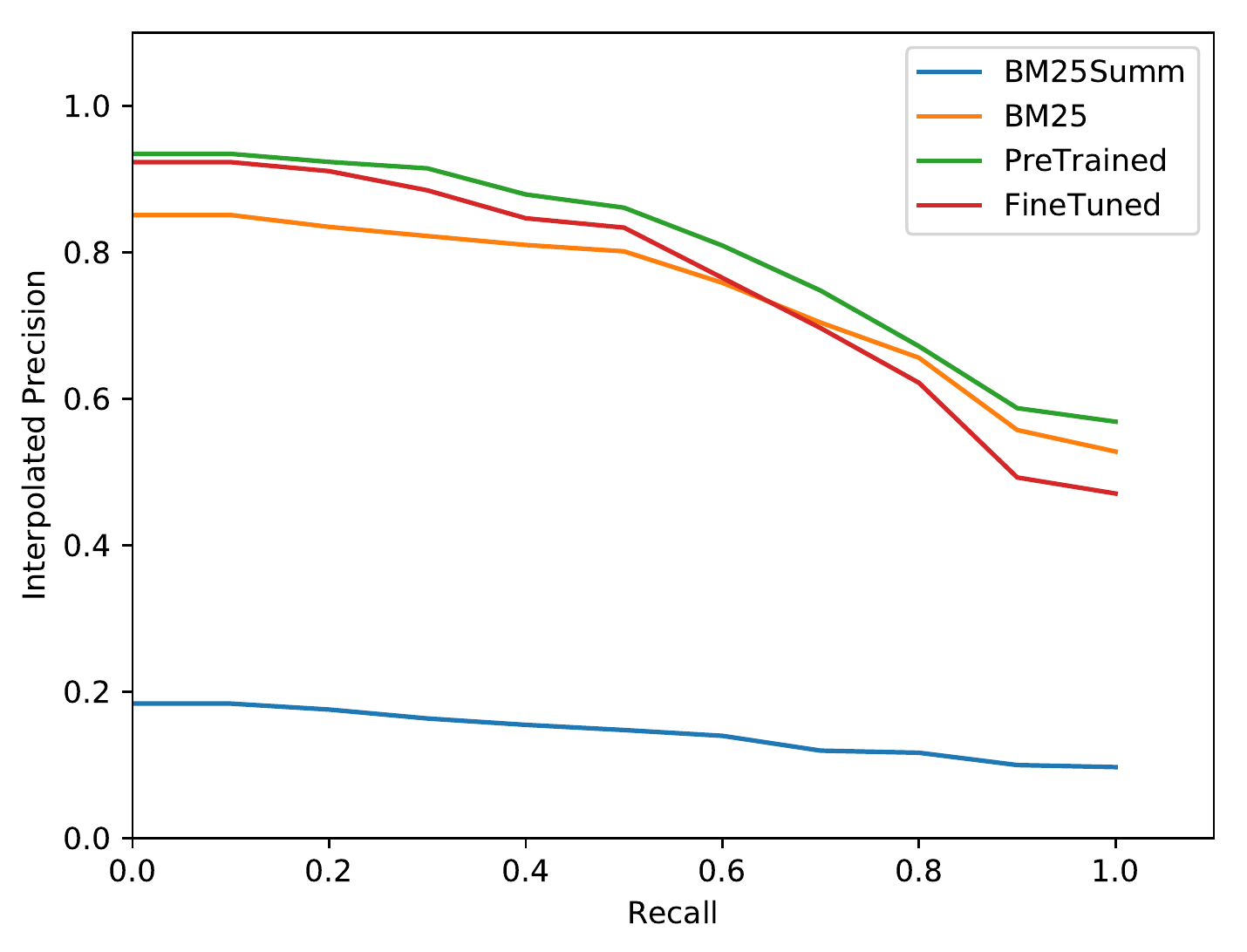}
  \caption{Precision-Recall curves}
  \label{01_pr_curve}
\end{figure}

We first observe that all of our models outperform the baseline of BM25 with summarized texts (model named 'BM25Summ'), and perform at a level comparable to BM25 on full texts. We interpret this part as a validation of our approach, and of the underlying assumption that summarization was a way to capture the information necessary to establish relevance, therefore bringing a positive answer to our research question RQ1.

The Precision-Recall curves in Figure~\ref{01_pr_curve}, show the skills of the introduced systems. It is noticeable that the \texttt{Fine-Tuned} system shows a consistent better performance at all levels of Recall.    

With regards to statistical significance, we computed the p-values of an OLS regression model with the target metric as the response variable, and a binary dummy (False for the Baseline, True for the tested model) as the independent variable. We made use of the python library Statsmodels~\cite{seabold2010statsmodels}. The p-value for the coefficient of the dummy indicates whether the observed difference in mean is significant or not. We use BM25 as the baseline, and observe a positive effect for both \texttt{PreTrained} and \texttt{FineTuned} models, while this effect is significant only for the \texttt{PreTrained} model.

We attribute this improvement to the knowledge gained during the additional pre-training. The amount of data available for our additional pre-training is orders of magnitude smaller that the material used for the initial pre-training, if we compare millions of words (our data) with tens of billions of words (Wikipedia, BookCorpus, etc.), so we consider this result as a positive answer to our research question RQ2.

\section{Statute Law Retrieval} \label{section_statute_retrieval}
\subsection{Experimental Setup}
Following the "Methods" section~\ref{section_methods}, we pre-process the collection by summarizing all texts. The final dataset is a collection of triplets made of query text, law article text and relevance judgment. We will fine-tune a pre-trained BERT model for a binary classification task, using the relevance labels from the labeled training dataset. We will also perform additional in-domain pre-training, and then fine-tune this model for the same classification task. In the results section~\ref{subsection_statute_law_retrieval_results}, the models appear under the names \texttt{FineTuned} and \texttt{PreTrained}.

For the in-domain pre-training, we used a collection of english translations of court decisions from the Japanese Supreme Court, scraped from their website\footnote{\url{http://www.courts.go.jp/app/hanrei\_en/search}}. This is a rather small dataset of 1500 texts, for a total of 230000 tokens. The training will start with a pre-trained BERT model (bert-base-uncased). We run the pre-training on 1 GPU nVidia Tesla P40, with 24GB onboard RAM during 2 hours, or 70000 iteration steps. We export this model for fine-tuning of 2 hours for 4 epochs.

For both models, after the fine-tuning step is finalized, the trained model computes the pairwise relevance score for each pair of query text and law article text in the test dataset, the score for the positive class is used to rank the law articles for each query.

\subsection{Baselines}
We use the SOTA from COLIEE 2018 as the baseline, namely \texttt{UB3}~\cite{yoshioka2018overview}. This system is based on Terrier\footnote{\url{http://terrier.org/}} and uses TagCrowd\footnote{\url{https://tagcrowd.com/}} for query reduction by extracting keywords.

\subsection{Results and Analysis} \label{subsection_statute_law_retrieval_results}
We report less metrics than for the previous task. As 98\% of queries have 1 or 2 relevant articles, we do not report Precision for ranks higher than 1. Results from different systems are compiled in Table~\ref{03_results_all}. 
\begin{table}[ht]
\centering
\begin{tabularx}{\linewidth}{|X|rrr|r|r|}
    \hline
    Name        & R@5  & R@10 & R@30  & MAP   & P@1 \\
    \hline
    UB3         & 0.7978 & 0.8539  & 0.9551 & 0.7988  &  \textit{unknown}   \\
    \hline
    FineTuned    & \textbf{0.9010} & \textbf{0.9203} & \textbf{0.9686} & 0.8246   & 0.7971 \\
    PreTrained    & 0.8913 & 0.9130 & 0.9444 & \textbf{0.8321}   & \textbf{0.8261} \\
    \hline
\end{tabularx}
\caption{Table of Results for All Systems}
\label{03_results_all}
\end{table}

Our work overperforms the baseline significantly. When focusing on only the models we introduce, we observe no significant differences in the evaluation metrics, with the exception of P@1 where the model with additional pre-training significantly improves over the "off the shelf" model. 

In this context, the additional pre-training yield a better performance for some metrics, but not others. The observed significant improvement on the P@1 metric is of high interest in the setting of a Question Answering task. We will consider this as a conditional positive answer to our research question RQ2, as the improvement might realize only on some specific metrics. 

\section{Conclusion} \label{section_conclusion}
We have demonstrated how models based on Generalized Language Models could perform in the context of Legal Information Retrieval, in the presence of limited data. We introduce a document summarization step in order to accomodate the sequence length limitations of BERT. In this setting, our system significantly improves over a BM25 system operating on full text.

We introduced an additional step of pre-training for existing models, which provided significant improvement in the task with the largest amount of training material. We leave for future work the transfer of this method to other specific legal domains. 

We consider for future work the possibilities of other new training tasks, considering Masked Language Modeling and Next Sentence Prediction as the way to establish a comprehension at semantic level, while other tasks would contribute to learn the deeper knowledge needed to achieve higher performance on the retrieval tasks.

\section*{Use of COLIEE Data}
We present these research findings based on the COLIEE Dataset for the Legal Case Retrieval Task, in accordance with the "MEMORANDUM ON PERMISSION TO USE ICAIL 2019 PARTICIPANT DATA COLLECTION". While we use some competition models as baseline, we do not claim that this paper is an entry to the official COLIEE competition. In the "Evaluation Metrics" section~\ref{subsection_evaluation_metrics}, we elaborate on how the metrics used for competition ranking differ from the metrics we present. We refer to the proceedings of ICAIL 2019~\cite{DBLP:conf/icail/2019} for further reading.

\bibliographystyle{ios1}
\bibliography{paper}

\end{document}